\documentclass{article}

\usepackage{PRIMEarxiv}

\usepackage[utf8]{inputenc} 
\usepackage[T1]{fontenc}    
\usepackage{hyperref}       
\usepackage{url}            
\usepackage{booktabs}       
\usepackage{amsfonts}       
\usepackage{nicefrac}       
\usepackage{microtype}      
\usepackage{lipsum}
\usepackage{fancyhdr}       
\usepackage{graphicx}       
\graphicspath{{media/}}     
\usepackage{cite}
\usepackage{amsmath,amssymb,amsfonts}
\usepackage{algorithmic}
\usepackage{graphicx}
\usepackage{textcomp}
\usepackage{mathrsfs}
\usepackage{subfig}
\usepackage{rsfso}
\usepackage{breqn}
\usepackage{caption}
\usepackage[square,sort,comma,numbers]{natbib}

\title{Kinematics and Dynamics Modeling of 7 Degrees of Freedom Human Lower Limb Using Dual Quaternions Algebra}

\author{
  Zineb BENHMIDOUCH, Saad MOUFID, Aissam AIT OMAR\\
  \texttt{z.benhmidouch@gmail.com} \\}

\begin{document}
\maketitle

\begin{abstract}
Denavit and Hartenberg-based methods, such as Cardan, Fick, and Euler angles, describe the position and orientation of an end-effector in three-dimensional (3D) space. However, these methods have a significant drawback as they impose a well-defined rotation order, which can lead to the generation of unrealistic human postures in joint space. To address this issue, dual quaternions can be used for homogeneous transformations. Quaternions are known for their computational efficiency in representing rotations, but they cannot handle translations in 3D space. Dual numbers extend quaternions to dual quaternions, which can manage both rotations and translations. This paper exploits dual quaternion theory to provide a fast and accurate solution for the forward and inverse kinematics and the recursive Newton-Euler dynamics algorithm for a 7-degree-of-freedom (DOF) human lower limb in 3D space.
\end{abstract}

\keywords{Dual numbers \and  Quaternions \and Dual quaternions algebra \and Artificial neural networks}

\section{Introduction}

Compared to classical methods such as Cardan, Fick, and Euler angles, which are based on homogeneous transformation, dual quaternions ~\cite{b1} offer an advantageous representation of rigid transformations in 3D space in many aspects. Dual quaternions require less computer memory, using only 8 elements to describe a rotation combined with translation of a rigid body in 3D space, whereas classical methods require 12 elements. Consequently, homogeneous transformation methods demand more storage and result in higher computational times due to the nonlinearity of the end-effector coordinates. Furthermore, these classical methods impose a well-defined rotation order. In contrast, the dual quaternion method reduces the number of mathematical operations, thus minimizing computational costs ~\cite{b2}. Additionally, the dual quaternion method helps to avoid discontinuities and singularities that can arise from the Euler angle representation, which is based on cylindrical polar coordinates for representing the motion of a rigid body in 3D space ~\cite{b3}.
\\
Furthermore, the dual quaternion method helps to avoid the gimbal lock phenomenon. Gimbal lock is a loss of one degree of freedom (1-DOF) in 3D space that occurs when using Euler angles  ~\cite{b4}. For example, suppose a rigid body rotates in 3D space in the order Z, Y, and X, and the angle of rotation about the Y-axis is 90°. In this case, the first and second rotations are performed correctly, but the rotation around the X-axis becomes impossible because the three rotation axes become coplanar.
\\

Dual quaternion algebra has been highlighted in numerous works, including the dynamics modeling of a mobile manipulator ~\cite{b5}, stabilization of rigid body motion, multiple body interactions ~\cite{b6}, inverse kinematic study of 6-DOF robot arms, and tracking control ~\cite{b7,b8}. For instance, Valverde et al. ~\cite{b9} presented a serial manipulator dynamics model using the recursive Newton-Euler method based on dual quaternions. Similarly, Silva et al. ~\cite{b10} employed the recursive Newton-Euler method and Gauss’s Principle of Least Constraint, both based on dual quaternions, to describe the relationships between joint velocities, forces, and torque variables of mobile manipulators
\\

In this paper, the dual quaternion-based theory is applied to the kinematics and dynamics study of the 7-DOF human lower limbs in 3D space. Subsequently, the artificial neural networks method is used to solve the inverse kinematics problem. The efficiency of the artificial neural networks method is verified using the jerk energy criteria. The rest of this paper is organized as follows: Section \ref{Preliminary theory} provides a brief mathematical background on dual quaternions algebra. Section \ref{Forward kinematics for lower limbs} elaborates on the forward kinematics of the human lower limb in 3D space using dual quaternions. Section \ref{Inverse kinematics for lower limbs} focuses on the application of the artificial neural network method to solve the inverse kinematics of the lower limb. In Section \ref{Dynamics modeling of lower limb}, the dynamical model of the lower limb using dual quaternions based on a recursive Newton-Euler method is developed. Finally, in Section \ref{Simulation results}, the simulation results are discussed.

\section{PRELIMINARY THEORY}

\label{Preliminary theory}
This section provides a theory background on dual quaternions required to describe rigid body motion in 3D-space. The rotation matrix $\mathbb{R}^{3\times3}$ is an orthogonal matrix defined by the Lie group $SO(3)\subset O(3)$, as follows:
\begin{equation}\label{eq:3}
    SO(3)=\left\{R \in \mathbb{R}^{3\times3}, R^TR=I_3, det (R)=\pm 1\right\}
\end{equation}
Where the orthogonal groups $O(3)$ and $SO(3)$ consist of the endomorphisms that preserve the euclidean norm and the normal direct isometries of $O(3)$, respectively. Thus, $SO(3)$ group preserves the orientation of the space \cite{b11}.
\\
The Euler angles convention is a common method to represent the special orthogonal group $SO(3)$. However, it exhibits a singularity at $\pm(n+1)\frac{\pi}{2}$, for all $n \in \mathbb{N}$. Furthermore, in certain configurations, there are infinite possible solutions for the sequence of rotations about the three axes when mapping from the special orthogonal group $SO(3)$ to Euler angles $\xi$ ~\cite{b12}.
\\
Euler’s Theorem on the axis of a 3D-space rotation states that if $R \in SO(3)$, then there is a non-zero vector $u$ that satisfies $Ru = u$. This means that the relative rotation of any two reference frames is equivalent to a single rotation about a given vector $u$ with a given rotation angle. Therefore, the orientation of a rigid body in 3D space can be described by a quaternion \cite{b13}:
\begin{equation}
    q= q_0+q_1i+q_2j+q_3k = \cos(\frac{\alpha}{2})+ \sin(\frac{\alpha}{2}) v
\end{equation} 
where $v$ and $\alpha$ are the rotation unit vector and angle, respectively.
\\
Multiplication of the quaternion components is done by using the following rules:
\begin{equation}
\begin{array}{c}
i^2=j^2=k^2=ijk=-1\\
ij=-ji=k\\
jk=-kj=i\\
ki=-ik=j\end{array}
\end{equation}
Let a quaternion $q$ represent the rotation of the vector $\vec{p}\in \mathbb{R}^3$ in 3D-space, thus the adjoint transformation is given by:
\begin{equation}\label{eq:8}
    \vec{p'}=Ad(q)\vec{p}=q\vec{p}q^*
\end{equation}
where $\vec{p'}\in \mathbb{R}^3$ is the resulting vector.
\\
Quaternions can only handle rotations of a rigid body around a fixed point. To also account for translations, the concept of dual numbers will be utilized. Introduced by Clifford in 1873 \cite{b14}, dual numbers, also known as duplex numbers, constitute a powerful algebraic structure for kinematic and dynamic analysis of spatial mechanisms. Dual numbers take the form $a + \epsilon b \in \mathbb{D}$, where $\epsilon$ is a new element termed the nilpotent number, satisfying $\epsilon^2 = \epsilon^3 = \ldots = \epsilon^n = 0$, with $n \in \mathbb{N}$. Dual quaternions extend quaternions using the concept of dual numbers, expressed as follows:
$$D=q_0+\epsilon q_1$$
Where $q_0$ and $q_1$ are quaternions denote the primary and dual parts of $D$, and $\epsilon$ represents the dual unit. The conjugate of $D$ is defined as:
\begin{equation}D^*=q_0^*-\epsilon q_1^*\end{equation}
Therefore, the dual quaternions are used to describe a pure rotation of a rigid body in space as follows:
\begin{equation}
    D_R=c(\frac{\alpha}{2})+s(\frac{\alpha}{2}) v
\end{equation}
where $\alpha$ and $v$ represent the rotation angle and unit vector, respectively. Hestenes \cite{b15}, shows that the derivative of the rotation dual quaternion is given by the following equation:
\begin{equation}
    \dot{D}_R=\frac{1}{2}D_R\vec{\Omega}_R
\end{equation}
Where $\vec{\Omega}_R$ denotes the dual quaternion of the rotation velocity. Further, the derivative of the conjugate of the rotation dual quaternion is given by:
\begin{equation}\label{eq:11}
    \dot{D}_R^*=-\frac{1}{2}\vec{\Omega}_RD_R^*
\end{equation}
\\
Furthermore, a pure translation is described by the following form:
\begin{equation}\label{eq:12}
    D_T=1+\epsilon \frac{d}{2}t
\end{equation}
Where $t$ and $d$ represent the unitary axis coordinates and the distance of the translation, respectively. Also, the translation conjugate is equal to itself. Whence, the velocity of the translation is given by:
\begin{equation}
   \dot{D}_T=\epsilon \frac{\dot{d}}{2}t+\epsilon \frac{d}{2}\dot{t}
\end{equation}
\\
The dual quaternion of the position and the velocity of a point $P$ are as follows:
\begin{equation}
\begin{array}{c}
P=1+\epsilon p\\
\dot{P}=\epsilon \dot{p}
\end{array}
\end{equation}
where $p$ is the coordinates of $P$ in 3D-space. 

\section{Forward kinematics for lower limbs}
\label{Forward kinematics for lower limbs}
\begin{figure}
    \centering
    \includegraphics[width=0.4\textwidth]{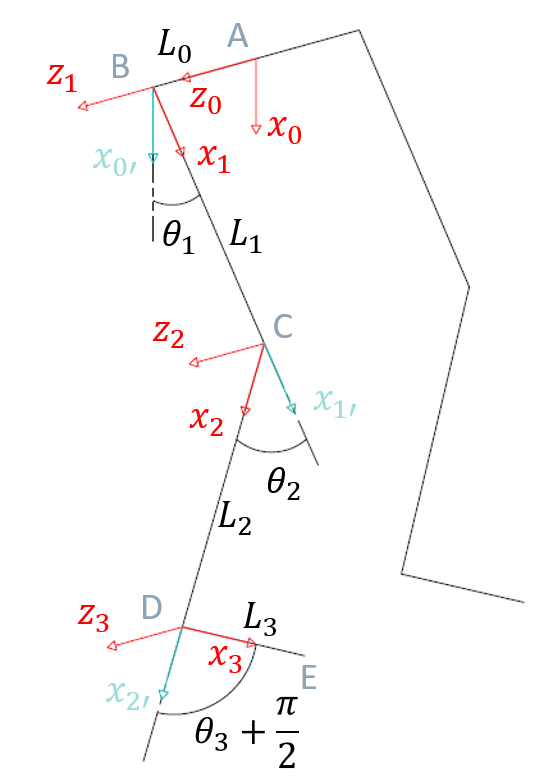}
    \centering
        \caption{Physiological lower limbs diagram}
    \label{fig:1}
\end{figure}

In this section, the Forward Kinematics (FK) of the lower limbs, depicted in Figure \ref{fig:1}, using dual quaternions is established. FK involves computing the positions and orientations of the end-effector in task space from the axes and angles of the joint rotations. The lower limb is decomposed into four segments: the pelvis, thigh, leg, and foot, connected by three joint groups. These include the hip, which rotates about three perpendicular axes; the knee, which moves solely about the z-axis; and the ankle, permitting movement in three planes. Therefore, the degrees of freedom (DOF) of the lower limbs total 7 \cite{b16}. Consequently, the position of the end-effector relative to the reference frame $\mathscr{R}3$, denoted as $P{E/3}$, can be expressed as:
\begin{equation}\label{eq:21}
P_{E/3}=Ad(r_3)Ad(t_{23})Ad(r_2)Ad(t_{12})Ad(r_1)Ad(t_{01})P_{E/0} 
    = r_3t_{23}r_2t_{12}r_1t_{01}P_{E/0}t_{01}r_1^*t_{12}r_2^*t_{23}r_3^*
\end{equation}
Where the numbers $1, ~2$ and $ 3$ stand for hip, knee and ankle, respectively. $P_{E/3}=1+\epsilon \frac{1}{2}\left(\begin{array}{ccc} L_3 & 0 & 0 \end{array}\right)^T$ and $P_{E/0}$ are the end-effector position relative to reference frame $\mathscr{R}_3$ and $\mathscr{R}_0$, respectively.\\
\begin{equation}
\label{eq:100}
    \begin{array}{c}
         r_1=c(\frac{\theta_1}{2})+n_1s(\frac{\theta_1}{2})  \\
         r_2=c(\frac{\theta_2}{2})+n_2s(\frac{\theta_2}{2})  \\
         r_3=c(\frac{\theta_3}{2}+\frac{\pi}{2})+n_3s(\frac{\theta_3}{2}+\frac{\pi}{2})
    \end{array}
\end{equation}\\
Equations (\ref{eq:100}) represent the thigh rotation around the pelvis, the leg rotation around the thigh, the foot rotation around the leg, respectively. where $n_1$, $n_2$ and $n_3$ are the rotation axis given by:
\begin{equation}
\label{eq:101}
    \begin{array}{c}
         n_1=\left(\begin{array}{ccc} n_{1x} & n_{1y} & n_{1z}\end{array}\right)^T  \\
         n_2=\left(\begin{array}{ccc} 0 & 0 & 1\end{array}\right)^T \\
         n_3=\left(\begin{array}{ccc} n_{3x} & n_{3y} & n_{3z}\end{array}\right)^T
    \end{array}
\end{equation}
The translation dual quaternions representing the offset between the pelvis center and the hip, the offset between the hip and the knee, and the offset between the knee and the ankle, respectively, are given by:
\begin{equation}
\label{eq:102}
    \begin{array}{c}
         t_{01}=1+\epsilon \frac{1}{2}\left(\begin{array}{ccc} 0 & 0 & L_0 \end{array}\right)^T  \\
         t_{12}=1+\epsilon \frac{1}{2}\left(\begin{array}{ccc} L_1 & 0 & 0 \end{array}\right)^T  \\
         t_{23}=1+\epsilon \frac{1}{2}\left(\begin{array}{ccc} L_2 & 0 & 0 \end{array}\right)^T
    \end{array}
\end{equation}
Therefore, the forward kinematics can be obtained using the following equation:
\begin{equation}\label{eq:22}
    P_{E/0}=t_{01}r_1^*t_{12}r_2^*t_{23}r_3^*P_{E/3}r_3t_{23}r_2t_{12}r_1t_{01}
\end{equation}
From Equation (\ref{eq:21}):
\begin{equation}
    P_{E/3}=r_3P_{E/2'} r_3^*
\end{equation}
Where $P_{E/2'}$ is the coordinates of the end-effector relative to reference frame $\mathscr{R}_2'$. Thereafter, the end-effector velocity relative to reference frame $\mathscr{R}_3$ is given by the following expression:
\begin{equation}
    \dot{P}_{E/3}=\dot{R}_3P_{E/2'} R_3^*+R_3\dot{P}_{E/2'} R_3^*+R_3P_{E/2'} \dot{R}_3^*
\end{equation}
Using Equations (\ref{eq:11}) and  (\ref{eq:12}), $\dot{P}_{E/3}$ is given as follows:
\begin{equation}
    \dot{P}_{E/3}=R_3\left(\frac{1}{2}\left(\Omega_3P_{E/2'}-P_{E/2'}\Omega_3\right)+\dot{P}_{E/2'}\right)R_3^*
\end{equation}
Where $\Omega_3$ is the rotation vector velocity of the foot around the leg. Hence:
\begin{equation}
    \dot{P}_{E/2'}=\dot{P}_{E/2}
\end{equation}
In the same way, the end-effector velocities relative to reference frame $\mathscr{R}_2$ and $\mathscr{R}_1$ are given as follows:
\begin{equation}
     \dot{P}_{E/2}=R_2\left(\frac{1}{2}\left(\Omega_2P_{E/1'}-P_{E/1'}\Omega_2\right)+\dot{P}_{E/1'}\right)R_2^*
\end{equation}
\begin{equation}
     \dot{P}_{E/1}=R_1\left(\frac{1}{2}\left(\Omega_1P_{E/0'}-P_{E/0'}\Omega_1\right)+\dot{P}_{E/0}\right)R_1^*
\end{equation}
Where $\Omega_2$ and $\Omega_1$ are the rotation vector velocities of the leg around the thigh and the thigh around the pelvis, respectively.
Therefore, the final expression of the end-effector velocity relative to reference frame $\mathscr{R}_0$ is given as follows:
\begin{equation}\label{equation20}
    \dot{P}_{E/0}=\frac{1}{2}\left(P_{E/0}\Omega_1-\Omega_1P_{E/0}\right) +\frac{1}{2}R_1^*\left(P_{E/1'}\Omega_2-\Omega_2P_{E/1'}\right)R_1+\frac{1}{2}R_1^*R_2^*\left(P_{E/2'}\Omega_3-\Omega_3P_{E/2'}\right)R_2R_1
\end{equation}

\section{Inverse kinematics for lower limbs}
\label{Inverse kinematics for lower limbs}
The Inverse Kinematics (IK) problem determines the joint angles required to achieve a desired end-effector position and velocity, facilitating efficient control of lower limb motion. However, the lower limb's excessive degrees of freedom (DOF) relative to its spatial constraints lead to redundancy issues. Consequently, solving the IK problem for the lower limb is complex, as it typically results in infinitely many solutions in 3D space. To address this challenge, a neural network method will be proposed in the following sections to manage redundancy and determine an optimal solution.
\\
Thereafter, the neural network is trained using data obtained from Equations (\ref{eq:22}) and (\ref{equation20}) to learn the angular values of joint configurations and rotation axes in the configuration space. In essence, this method enables the solution of the nonlinear IK problem by identifying the joint configurations $\theta_1, \theta_2, \theta_3$ and rotation axes $n_1 = (n_{1x}, n_{1y}, n_{1z})^T$ and $n_3 = (n_{3x}, n_{3y}, n_{3z})^T$ that correspond to the predefined end-effector position $(x_d, y_d, z_d)^T$ and velocity $(\dot{x}_d, \dot{y}_d, \dot{z}_d)^T$. To ensure an efficient database creation, workspace constraints must be respected. Therefore, the rotation axis constraints are specified as follows:
\begin{equation}
    \bigg\{\begin{array}{c}
    n_{3x}^2+n_{3y}^2+n_{3z}^2=1\\
    n_{1x}^2+n_{1y}^2+n_{1z}^2=1\end{array}
\end{equation}
\\
Moreover, the hip and knee joints exhibit mechanical characteristics of spherical joints, enabling movement in three planes, whereas the knee joint forms a pivot joint. However, these movements are constrained by upper and lower limits as shown in Table \ref{tab:1}. Therefore, the following inequalities allow for the generation of a set of angular joint values while adhering to the range of motion constraints for the hip, knee, and ankle:

\begin{equation}
    \begin{array}{c}
    -30^{\circ} \leqslant arccos\left(\frac{P_{C/0}-P_{B/0}}{\vert P_{C/0}-P_{B/0} \vert}\cdot\vec{x_0}\right) \leqslant 120^{\circ}\\
    -20^{\circ} \leqslant arccos\left(\frac{P_{C/0}-P_{B/0}}{\vert P_{C/0}-P_{B/0} \vert}\cdot\vec{z_0}\right) \leqslant 45^{\circ}\\
    -50^{\circ} \leqslant arccos\left(\frac{P_{C/0}-P_{B/0}}{\vert P_{C/0}-P_{B/0} \vert}\cdot\vec{y_0}\right) \leqslant 40^{\circ}\\
   -150^{\circ} \leqslant \theta_2 \leqslant 0^{\circ}\\
    -40^{\circ} \leqslant arccos\left(\frac{P_{E/0}-P_{D/0}}{\vert P_{E/0}-P_{D/0} \vert}\cdot\vec{x_0}\right) \leqslant 20^{\circ}\\
    -35^{\circ} \leqslant arccos\left(\frac{P_{E/0}-P_{D/0}}{\vert P_{E/0}-P_{D/0} \vert}\cdot\vec{z_0}\right) \leqslant 30^{\circ}\\
    -35^{\circ} \leqslant arccos\left(\frac{P_{E/0}-P_{D/0}}{\vert P_{E/0}-P_{D/0} \vert}\cdot\vec{y_0}\right) \leqslant 20^{\circ}\\
   \end{array} 
\end{equation}

\begin{table}[h]
    \centering
    \caption{Range of motion of lower limb}
    \begin{tabular}{|c|c|c|}
      \hline
        Joint & Movements & Range of motion\\
         \hline
        Hip & Flexion/Extension & From $-30^{\circ}$ to $120^{\circ}$ \\
      
         & Adduction/Abduction  & From $-20^{\circ}$ to $45^{\circ}$\\
          
         & Medial/Lateral rotation  & From $-50^{\circ}$ to $40^{\circ}$\\
         \hline
         Knee & Flexion/Extension & From $-150^{\circ}$ to $0^{\circ}$ \\
           \hline
         Ankle & Plantarflexion/Dorsiflexion & From $-40^{\circ}$ to $20^{\circ}$ \\
           & Pronation/External rotation  & From $-35^{\circ}$ to $30^{\circ}$\\
          
         & Inversion/Eversion  & From $-35^{\circ}$ to $20^{\circ}$\\
         \hline
    \end{tabular}
    \label{tab:1}
\end{table}

\section{Dynamics modeling of lower limb}
\label{Dynamics modeling of lower limb}
\begin{figure}
    \centering
    \includegraphics[width=0.35\textwidth]{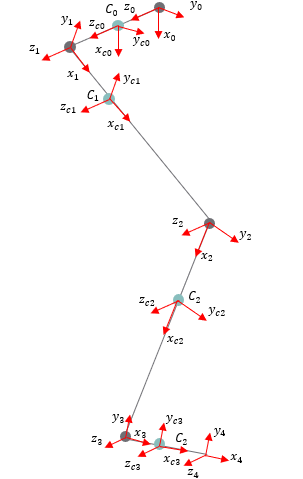}
    \centering
        \caption{Lower limb with 7-DOF}
    \label{fig:18}
\end{figure}

This section focuses on the dynamic description of the lower limb shown in Figure \ref{fig:18} using the Dual Quaternion-based recursive Newton-Euler method. This method involves calculating the velocities and accelerations of the center of mass of each link, known as twists, based on the positions, velocities, and accelerations of the lower limb configuration. These calculations adhere to the Newton-Euler propagation law. Subsequently, the wrenches, representing forces and moments acting on each link in 3D space, are derived starting from the wrenches applied to the end effector.
\subsection{ Forward Recursion}
The forward recursion aims to obtain the twist, which includes the angular and linear velocities of each link's center of mass, along with its first derivative. Let $\xi_{0,c_0}^{c_0}$ denote the twist of the center of mass $C_0$ with respect to frame $\mathscr{R}{0}$, expressed in frame $\mathscr{R}{c_0}$. It is represented by the following dual quaternion:
\begin{equation}
   \xi_{0,c_0}^{c_0}= \omega_{0,c_0}^{c_0}+\epsilon v_{0,c_0}^{c_0}=0
\end{equation}
where $\omega_{0,c_0}^{c_0}$ and $v_{0,c_0}^{c_0}$  are, respectively, the angular and linear velocities. In general, the twist $ \xi_{0,c_i}^{c_i},~~i\in[1,3]$ expressed in reference frame $\mathscr{R}_{c_i}$ that yields the motion of $\mathscr{R}_{c_i}$ with respect to frame $\mathscr{R}_{0}$, is given by:
\begin{equation}\label{eq:152}
    \xi_{0,c_i}^{c_i}=\xi_{0,i}^{c_i}+\xi_{i,c_i}^{c_i}
\end{equation}
where $\xi_{0,i}^{c_i}$ and $\xi_{i,c_i}^{c_i}$ represent the twists related to the motion of the $i$-th joint and the $i$-th link's center of mass, respectively. The twist of the $i$-th link's center of mass with respect to the reference frame $\mathscr{R}_{0}$ depends on the twist provided by its joint and on the twist of the $(i-1)$-th relative to the $i$-th joint because they are physically attached. Therefore, using Equation (\ref{eq:8}), Equation (\ref{eq:152}) can be expressed as follows:
\begin{equation}
    \xi_{0,c_i}^{c_i}  =\xi_{0,i}^{c_i}+\xi_{i,c_i}^{c_i}= Ad(x_{c_{i-1}}^{c_i})(\xi_{0,c_{i-1}}^{c_{i-1}}+\xi_{c_{i-1},i}^{c_{i-1}})+Ad(x_{i}^{c_i})\xi_{i,c_i}^{i}
\end{equation}
where $x_{c_{i-1}}^{c_i}=r_{c_{i-1}}^{c_i}+\frac{1}{2}\epsilon r_{c_{i-1}}^{c_i}p_{c_{i-1}}^{c_i}$ and $x_{i}^{c_i}=r_{i}^{c_i}+\frac{1}{2}\epsilon r_{i}^{c_i}p_{i}^{c_i}$ are units dual quaternions where $r$ and $p$ denote rotation and  position in 3D-space, respectively.As well as, $\xi_{c_{i-1},i}^{c_{i-1}}=0$ because $x_{c_{i-1},i}^{c_{i-1}}$ is constant. Thus,
\begin{equation}\label{eq:153}
    \xi_{0,c_i}^{c_i}  = Ad(x_{c_{i-1}}^{c_i})\xi_{0,c_{i-1}}^{c_{i-1}}+Ad(x_{i}^{c_i})\xi_{i,c_i}^{i}
\end{equation}
where, 
\begin{equation}
\xi_{i,c_i}^{i}=\omega_{i,c_i}^{i}+\frac{1}{2}\epsilon(\dot{P}_{i,c_i}^{i}+\omega_{i,c_i}^{i}\times P_{i,c_i}^{i})
\end{equation}
is the twist of reference frame $\mathscr{R}_{c_i}$ with respect to reference frame $\mathscr{R}_{i}$, expressed in
frame $\mathscr{R}_{i}$.\\
where,
\begin{equation}
    P_{i,c_i}^{i}\times\omega_{i,c_i}^{i}=\frac{P_{i,c_i}^{i}\omega_{i,c_i}^{i}-\omega_{i,c_i}^{i} P_{i,c_i}^{i}}{2}
\end{equation}
Therefore, the time derivative of Equation (\ref{eq:153}) is given as follows:
\begin{equation}
\begin{split}
     \dot{\xi}_{0,c_i}^{c_i} =Ad(x_{c_i-1}^{c_i})\dot{\xi}_{0,c_i-1}^{c_i-1}+Ad(x_{i}^{c_i})\dot{\xi}_{i,c_i}^{i}+\xi_{c_i,c_i-1}^{c_i}\times(Ad(x_{c_i-1}^{c_i})\xi_{0,c_i-1}^{c_i-1})
\end{split}
\end{equation}
\subsection{Backward Recursion}
Backward recursion involves calculating the required joint torques to achieve the predefined motion based on the dual quaternion wrenches acting at the end effector, along with the twists and their first time derivatives for each link's center of mass. The wrench at the foot's center of mass is given by:
\begin{equation}
    \zeta_{0,c_3}^{c_3}=f_{0,c_3}^{c_3}+m_3g^{c_3}+\epsilon\tau_{0,c_3}^{c_3}
\end{equation}
where $g^{c_3}$ is the force of gravity and $f_{0,c_3}^{c_3}$ is the force applied at the foot's center of mass. Moreover, $\tau_{0,c_3}^{c_3}$ is the torque about the the foot's center of mass results of its angular momentum variation, given by:
\begin{equation}
    \tau_{0,c_3}^{c_3}=\chi_3(I_3^{c_3})\mathcal{P}(\dot{\xi}_{0,c_3}^{c_3})+\mathcal{P}(\xi_{0,c_3}^{c_3})\times\chi_3(I_3^{c_3})\mathcal{P}(\xi_{0,c_3}^{c_3})
\end{equation}
where $\mathcal{P}$ return the primary component, $I_3^{c_3}=(i_x,i_y,i_z)\in \mathbb{H}^3$ is the quaternionic inertia tensor and $\chi_3$ is the operator defined as
\begin{equation}
    \chi_3(I)\mathcal{P}=<i_x,\mathcal{P}>i+<i_y,\mathcal{P}>j+<i_z,\mathcal{P}>k
\end{equation}
where $<.,.>:\mathbb{H}\rightarrow\mathbb{R},~~<a,b>=-\frac{ab+ba}{2}$ is the inner product. Using the adjoint transformation, the wrench $\zeta_{0,3}^{c_3}$ is given by:
\begin{equation}
    \zeta_{0,3}^{c_3}=Ad(x_{c_3}^{3})\zeta_{0,c_3}^{c_3}
\end{equation}
Thereafter, The wrench $\zeta_{0,i}^{c_i},~~i\in[1,2]$ of $i^{th}$ joint is given by:
\begin{equation}
    \zeta_{0,i}^{c_i}=Ad(x_{c_i}^{i})\zeta_{0,c_i}^{c_i}+Ad(x_{i+1}^{i})\zeta_{0,i+1}^{c_{i+1}}
\end{equation}
where $\zeta_{0,c_i}^{c_i}=f_{0,c_i}^{c_i}+m_ig^{c_i}+\epsilon\tau_{0,c_i}^{c_i}$ is the wrench at the $i^{th}$ link's center of mass.
\\
Extraction of $\tau_{c_i}^{c_i}$ components of the wrench $\zeta_{0,c_i}^{c_i}$ yield the dynamic equation of the lower limb that can be written in the following form:
\begin{equation}
    \tau=M(q)\ddot{q}+V(q,\dot{q})+G(q)
\end{equation}
where $M(q)$ is the mass matrix of the lower limb, $V(q,\dot{q})$ is the centrifugal and Coriolis vector, and $G(q)$ is the gravity force vector.
\begin{figure}
    \centering
    \includegraphics[width=0.6\textwidth]{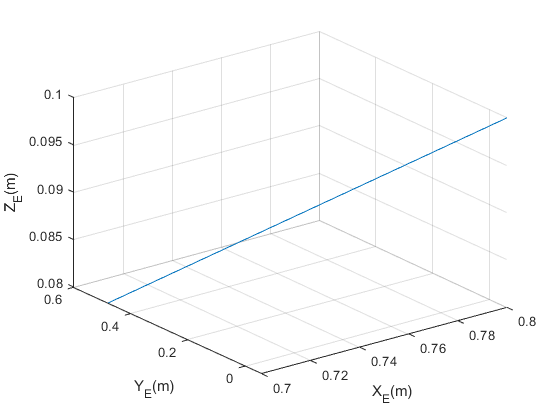}
        \caption{Trajectory planning for the lower limb in 3D-space}
    \label{fig:4}
\end{figure}
\begin{figure}
    \centering
    \includegraphics[width=0.8\textwidth]{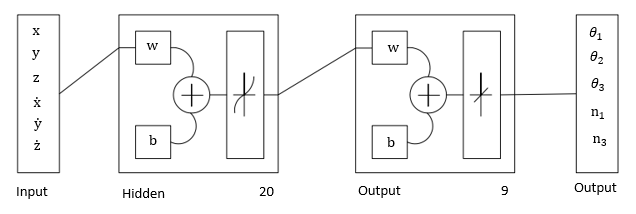}
        \caption{ANN algorithm}
    \label{fig:6}
\end{figure}
\begin{figure}%
    \centering
    \subfloat[\centering Rotation vectors coordinates]{{\includegraphics[width=7.5cm]{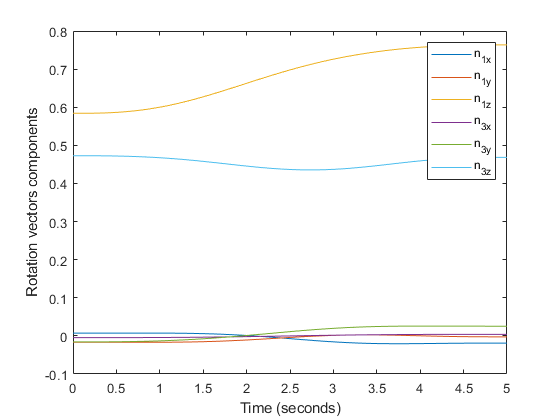} }}%
    \qquad
    \subfloat[\centering Articulation angle values]{{\includegraphics[width=7.5cm]{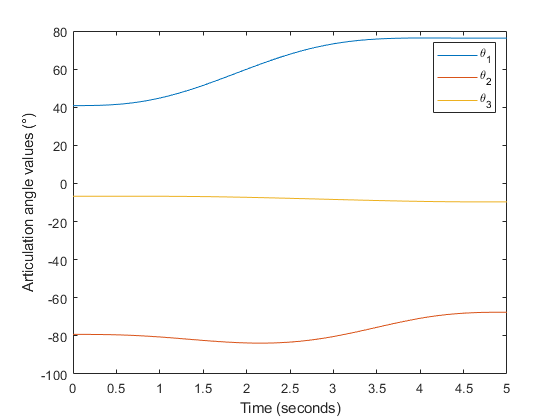} }}%
    \caption{Simulation results}%
    \label{fig:7}%
\end{figure}
\begin{figure}
    \centering
    \includegraphics[width=0.6\textwidth]{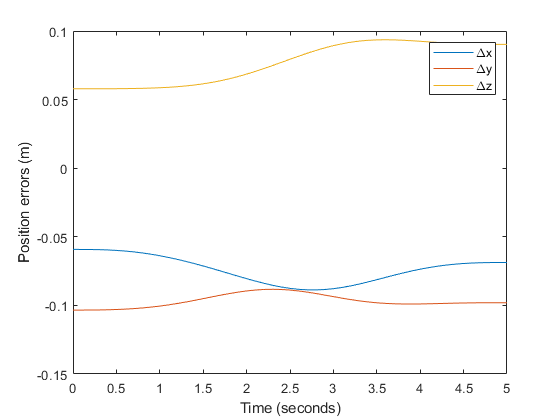}
    \caption{Position Error }
    \label{fig:8}
\end{figure}

\begin{figure}
    \centering
    \includegraphics[width=0.6\textwidth]{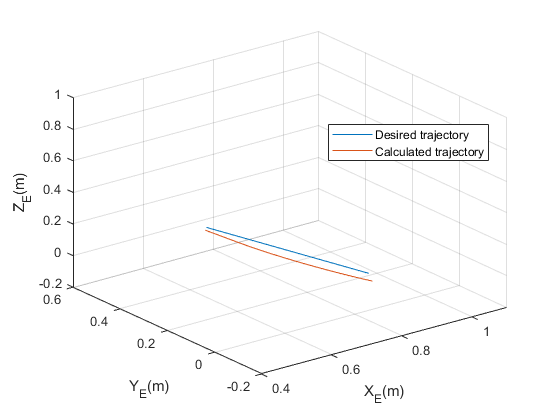}
    \caption{Desired and calculated trajectory in 3D-space}
    \label{fig:11}
\end{figure}

\begin{figure}
    \centering
    \includegraphics[width=0.6\textwidth]{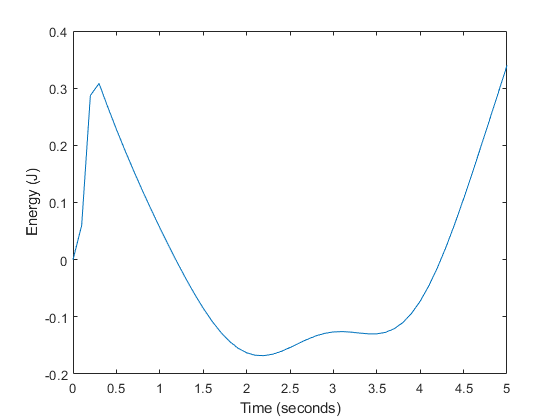}
        \caption{Energy of the motion}
    \label{fig:12}
\end{figure}
\section{Simulation results}
\label{Simulation results}
Initially, the set of waypoints required to generate a trajectory in task space is assumed to be available from the task planner. The trajectory is then planned using the minimum jerk criterion to ensure smooth acceleration of the joints, thereby reducing vibrations and avoiding resonance frequencies. For this simulation example, the initial and final position, velocity, and acceleration values are provided in Table \ref{tab:2}. The simulation result of the trajectory planning is depicted in Figure \ref{fig:4}.
\begin{table}
    \centering
    \caption{Initial and final position, velocity, acceleration of a motion values}
    \begin{tabular}{|c|c|c|}
      \hline
        &  Initial value & Final value \\
        \hline
         Position ($m$) &  $(0.8,-0.06,0.1)$ & $(0.7,0.48,0.08)$
            \\
           \hline
         Speed ($m/s$) &  0 & 0   \\
            \hline
         Acceleration ($m/s^2$) &  0 & 0 \\
         \hline
    \end{tabular}
    \label{tab:2}
\end{table}

Afterward, the inverse kinematics (IK) of the lower limb is computed using a multi-layer perceptron trained with the Levenberg-Marquardt backpropagation algorithm, utilizing a dataset of 400,000 samples. The network architecture is illustrated in Figure \ref{fig:6}, featuring a two-layer feed-forward structure comprising a hidden layer with 20 interconnected sigmoid neurons and an output layer with 9 linear neurons. The simulation results are presented in Figure \ref{fig:7}, achieving a computational time of 0.009 s. Additionally, Figure \ref{fig:8} illustrates the position errors along the x-axis, y-axis, and z-axis, with root mean square errors of 0.0741, 0.0970, and 0.0776, respectively. The negligible position error between the desired and calculated trajectories shown in Figure \ref{fig:11} confirms the method's accuracy.

Furthermore, to minimize energy expenditure and combat fatigue, humans ideally produce motion characterized by low energy output. This energy constraint on the angular joints corresponds to the third derivative, aligning with the minimum jerk criterion. Figure \ref{fig:12} demonstrates that the cumulative energy of $\theta_1$, $\theta_2$, and $\theta_3$ joints is less than 0.4 J.
\section{Conclusion}
The primary objective of this paper was to leverage dual quaternions algebra for describing the kinematics, encompassing position and orientation, as well as the dynamics modeling of an anthropomorphic leg in 3D-space, thereby circumventing the high computational costs associated with homogeneous transformation methods. To achieve this, artificial neural networks (ANN) were employed to solve the inverse kinematics (IK) problem while adhering to range of motion constraints, and the minimum energy criterion was applied to ensure realistic human posture. Additionally, the Newton-Euler recursive method based on dual quaternions was chosen for dynamics modeling to mitigate the complexities associated with geometric analyses.\\

\end{document}